\documentclass[letterpaper]{article}
\usepackage{authblk}
\usepackage{aaai}
\usepackage{times}
\usepackage{helvet}
\usepackage{courier}
\usepackage{gensymb}
\usepackage{natbib}
\usepackage[export]{adjustbox}
\DeclareMathOperator*{\argmax}{arg\,max}
\title{Smart Library: Identifying Books in a Library using Richly Supervised Deep Scene Text Reading}

\author[1]{Xiao Yang}
\author[2]{Dafang He}
\author[2]{Wenyi Huang}
\author[2]{Zihan Zhou}
\author[2]{Alex Ororbia}
\author[1]{Daniel Kifer}
\author[1,2]{C. Lee Giles}

\affil[1]{Department of Computer Science and Engineering, Pennsylvania State University}
\affil[2]{School of Information Science and Technology, Pennsylvania State University}

\begin{document}
\setlength{\abovedisplayskip}{10pt}
\setlength{\belowdisplayskip}{10pt}
\setlength\abovedisplayshortskip{10pt}
\setlength\belowdisplayshortskip{10pt}
\setlength{\bibsep}{0.3em}

\maketitle

\begin{abstract}
\begin{quote}
Physical library collections are valuable and long standing resources for knowledge and learning. However, managing books in a large bookshelf and finding books on it often leads to tedious manual work, especially for large book collections where books might be missing or misplaced. Recently, deep neural models, such as Convolutional Neural Networks (CNN) and Recurrent Neural Networks (RNN) have achieved great success for scene text detection and recognition. Motivated by these recent successes, we aim to investigate their viability in facilitating book management, a task that introduces further challenges including large amounts of cluttered scene text, distortion, and varied lighting conditions. In this paper, we present a library inventory building and retrieval system based on scene text reading methods. We specifically design our scene text recognition model using rich supervision to accelerate training and achieve state-of-the-art performance on several benchmark datasets. Our proposed system has the potential to greatly reduce the amount of human labor required in managing book inventories as well as the space needed to store book information.
\end{quote}
\end{abstract}

\vspace{-1em}
\section{Introduction}
Despite the increasing availability of digital books in the modern age of information, many people still favor reading physical books. The large libraries that house them require great amount of time and labor to manage inventories that number in the millions. Manually searching bookshelves is time-consuming and often not fruitful depending on how vague the search is. In this paper, we propose a deep neural network-based scene text reading system that can viably solve this problem.

\begin{figure}[h!]
	\vspace{-1em}
    \includegraphics[width=0.45\textwidth,center]{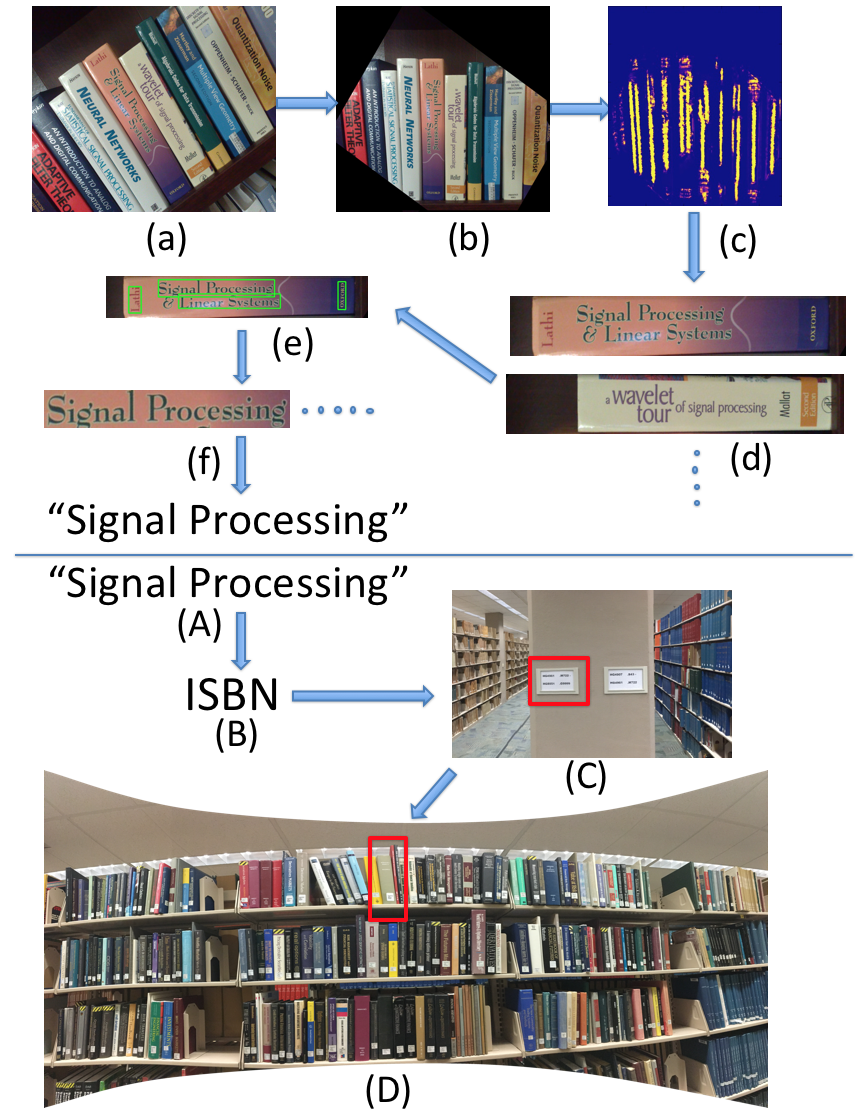}
    \caption{The pipeline of our system. (a)-(f) corresponds to building a book inventory while (A)-(D) corresponds to locating a book in a library. (a) Original image. (b) Rotated image based on estimation of dominant orientation. (c) Generated saliency image. (d) Segmented book spine. (e) Detected lines of words. (f) Cropped image patch for robust reading. We not only read titles, but also read other words on the book spine which provide rich book-identifying information. (A) Book title as query keywords. (B) Corresponding ISBN number. (C) Location of the stack the book belongs to. (D) Location of the book in the shelf.}
    \vspace{-1.5em}
    \label{fig:pipeline}
\end{figure}

Our work has two goals: 1) build a book inventory from only the images of bookshelves; 2) help users quickly locate the book they are looking for. Once the book is identified, additional stored information could also be retrieved, e.g. a short description of the book. The intent of this work is to make what was previously a tedious experience (i.e., searching books in bookshelves) much more user-friendly, especially for visually impaired people. In addition, our system also offers the potential for efficiently building a large database of book collections. The full pipeline of our system is summarized in Figure \ref{fig:pipeline}.

Our contributions are as follows:

\begin{itemize}
\item We build an end-to-end deep scene text reading system specifically designed for book spine reading and library inventory management. We demonstrate that the ``text information'' extracted by our deep neural network-based system alone can achieve good retrieval performance. This is essential, since other data, like digital images of all book covers in a collection, are not available to all users. To the best of our knowledge, using neural-based scene text reading to facilitate library inventory construction has not been explored before. 

\item For text localization in a library environment, we design a book spine segmentation method based on Hough transform \citep{duda1972use} and scene text saliency. A state-of-the-art text localization method is subsequently applied to extract multi-oriented text information on a segmented book spine.

\item For text recognition, we adopt a deep sequential labeling model based on convolutional and recurrent neural architectures. We propose using a per-timestep classification loss in tandem with a revised version of the Connectionist Temporal Classification (CTC) \citep{graves2006connectionist} loss, which accelerates the training process and improves the performance. Our model achieves state-of-the-art recognition performance on several benchmark datasets.
\end{itemize}


\vspace{-0.5em}
\section{Related Work}\label{sec:related_work} 
Previous work in book inventory management has typically focused on book spine detection for which many different algorithms have been proposed. A general framework of book spine extraction was proposed by \citet{quoc2009framework}, and several challenges were addressed, such as lighting condition and distortion of images. Systems have been also designed for reading book spines including that of \citet{chen2010building,nevetha2015automatic,lee2008matching,tsai2011combining}. In \citet{taira2003book}, a finite state automata model for book boundary detection, combined with a global model-based optimization algorithm, was formulated for detecting the angle and boundary of each book. \citet{chen2010building} designed an efficient Hough Transform \citep{duda1972use} based book boundary detector to be used with SURF \citep{bay2006surf} feature image matching in order to retrieve books in an inventory. \citet{tsai2011combining} proposed a hybrid method that combined a text-reading-based method with an image-matching-based method.

It is important to note that the performance of most existing approaches is limited by book spine segmentation and off-the-shelf OCR systems. Hand-craft features based book spine segmentation suffers from common image distortion and low contrast between books. Off-the-shelf OCR system, such as Tesseract \citep{smith2007overview}, perform poorly on image taken of natural scenes. Recently, scene text reading has gained a great deal of attention in the computer vision community and we argue that book spine recognition should be considered as an challenging scene text reading problem. In this paper, we present a deep neural network based system and demonstrate that scene text reading can be effectively applied to book information retrieval and book inventories management. Combined with other image processing techniques, such as the Hough Transform, our system achieves robust performance on book information retrieval.

\section{Text Localization}
\label{sec:localization}

In this section, we present our method for detecting text in images of library bookshelves.

\subsection{Book Spine Segmentation}
\label{spine_segment}
Book spine segmentation is a critical component of our system since each book is expected to be recognized, stored, and queried independently. Most existing methods rely exclusively on low-level methods for segmentation, such as Hough Transform. We, however, only use Hough Transform as a pre-processing step to extract the dominant direction of books in the image. The dominant direction is then used to rotate the entire image (Figure \ref{fig:pipeline}(b)).

After rotating the image, we apply a text/non-text CNN model\footnote{The convolutional parts of the architecture are shared with later text recognition modules.} trained on $32 \times 64$ color image patches to generate saliency maps of the rotated image.

The saliency images could be further used in the following ways: (1) extract the book title location, and (2) segment each book. As shown in Figure \ref{fig:pipeline}(c), the saliency images provide a clear segmentation of each book even when two neighbor book spines have the same color. We simply use a non-max suppression to find the segmenting point of each book along the vertical axis. As a result, we circumvent the need for book spine segmentation methods based on the Hough Transform or other low-level routines, which can be easily influenced by lighting conditions or low contrast between books. 

\subsection{Text Localization within Book Spine}
\label{spine_local}
A scene text detection algorithm is subsequently applied to each book spine our system obtains. This step detects words on the book spines, which provide detailed and useful information about the book. 

In the literature, scene text detection method generally falls into one of two categories: sliding-window based methods \citep{chen2004detecting} and region proposal based methods \citep{neumann2012real,huang2014robust}. In this work, a region based method is adopted. We first generate extreme regions as in \citet{neumann2012real}. The extreme region approach has two properties that make it suitable for text detection: (1) fast computation, and (2) high recall. The saliency maps generated by the CNN are then used to filter out non-text regions. After this, candidate text regions are retained and a multi-orientation text line grouping algorithm is applied to find different lines of text. Here, we follow the work of \citet{dafang2016aggre} by first constructing a character graph and then aligning character components into text lines. Low level features, such as perceptual divergence and aspect ratio, are used to find text components that belong to the same line.

Because of the multi-oriented nature of the text lines in book spines, we need to further decide whether a text line is upside down or not. To address this issue, we train a CNN classifier on $32 \times 96$ image patches. The binary classifier tells us whether we need to flip the text lines in order to provide the text recognition module with a correct sequence. During testing, we first resize the height of all the text lines to 32 pixels. Then, we slide a window of size $32 \times 96$ on the text line image patch to obtain a series of predictions. If the average probability of being upside down is greater than $0.7$, we will rotate the text line image patch by $180\degree$. If the probability is lower than $0.3$, we just keep the original text line. If the probability is in between, we will keep both orientations. This decision schema was devised so as to reduce the influence of false positives on retrieval task later. If we fail to retain the correct orientation, we might not be able to find the corresponding book. Finally, it is worth mentioning that we choose a $32 \times 96$ window because text detected in smaller windows (e.g. $32 \times 32$) might be less informative in terms of its orientation. For example, a window containing `O' or `OO' does not provide much information whether a line should be flipped or not.

\vspace{-0.8em}
\section{Text Recognition}
\label{sec:recognition}
In our system, book spine images are identified based on the recognized text, which are then used as keywords for indexing or searching a book database. During the querying process, our system only relies on text information without requiring the original spine images.

\subsection{Text Recognition via Sequence Labeling}
\label{seq_labeling}
In the text recognition step, a common approach is to first segment and recognize each character, then output a word-level prediction based on a language model or a combination of heuristic rules. However, these approaches are highly sensitive to various distortions in images, making character-level segmentation imperfect and sometimes even impossible. To bypass the segmentation step, our model is trained to recognize a sequence of characters simultaneously. Similar to \citet{he2016reading} and \citet{ShiBY15}, we adopt a hybrid approach that combines a CNN with an RNN, casting scene text recognition problem as a sequential labeling task. A CNN at the bottom learns features from the images which are then composed as feature sequences that are subsequently fed into an RNN for sequential labeling.


\begin{figure*}
    \centering
    \vspace{-1.5em}
    \includegraphics[width=0.8\textwidth]{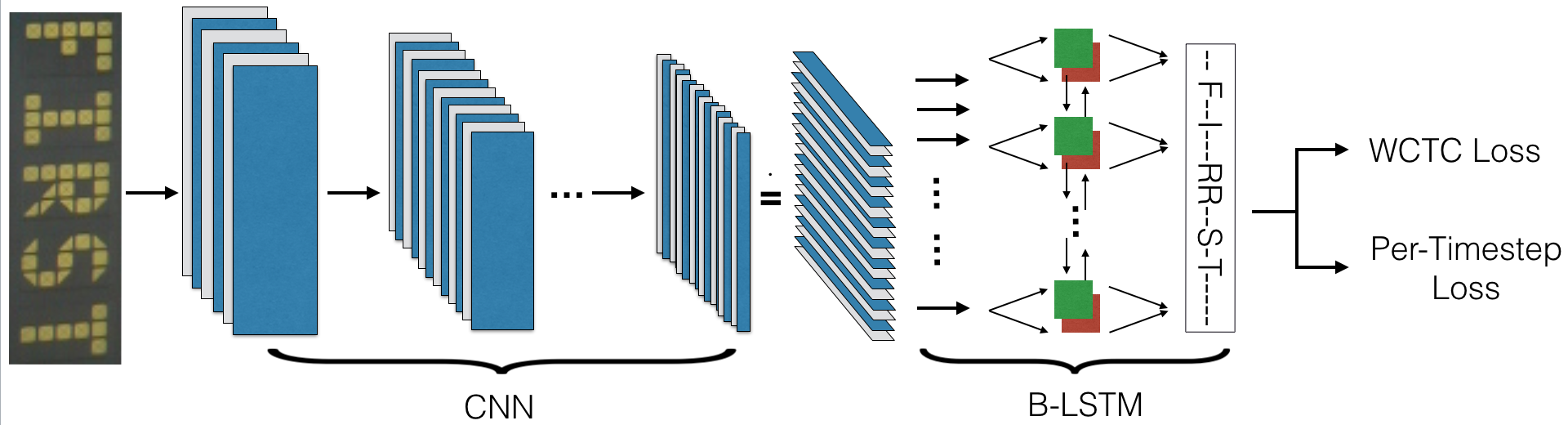}
    \vspace{-0.5em}
    \caption{The architecture of the proposed recognition model. The CNN element is similar to VGG16 \citep{simonyan2014very} except that the stride of the pooling layers are adjusted such that the last convolutional feature map has a height of $1$. Batch normalization \citep{ioffe2015batch} is added after each convolution layer to accelerate convergence.}
    \vspace{-1.5em}
    \label{fig:recognition_architecture}
\end{figure*}

Figure \ref{fig:recognition_architecture} summarizes the architecture of the proposed text recognition model. We first generate a sequence of deep CNN features $F = \{f_1, f_2, \cdots , f_T\}$ from an image $I$. To further exploit the interdependence among features, a bidirectional Long Short-Term Memory (B-LSTM) model \citep{hochreiter1997long} is stacked on top of generated sequential CNN features, yielding another sequence $X = \{x_1, x_2, \cdots , x_T\}$ as final outputs. Each one of the $x_i$ is normalized through a softmax function and can be interpreted as the emission of a specific label for a given time-step. On the other hand, the target word $Y$ can be viewed as a sequence of characters, $Y = \{y_1, y_2, \cdots , y_L\}$.

\subsection{The CTC Loss}
\label{ctc_loss}
Note that sequences $X$ and $Y$ have different lengths $T$ and $L$ respectively, which make it difficult to train our model in the absence of a per-timestep groundtruth. Following the work of \citet{graves2006connectionist}, we adopt CTC loss to allow an RNN to be trained for sequence labeling task without exact alignment.

CTC loss is the negative log likelihood of outputting a target word $Y$ given an input sequence $X$:
\vspace{-0.5em}
\begin{align}
	\text{CTC}(X) = - \text{log} P (Y | X)
\end{align}
\vspace{-1.5em}

\noindent
Suppose there is an alignment $a$ which provides a per-timestep prediction (whether blank or non-blank labels) for $X$, and a mapping function $B$ which would remove repeated labels and blanks from $a$. For instance, $(-, a, a, -, -, b)$ would be mapped as $(a,b)$ (using $-$ to denote blank label). Then, $P (Y | X)$ can be calculated by summing all possible alignments $a$ that can be mapped to $Y$:
\vspace{-0.5em}
\begin{align}
	P (Y | X) = \sum_{a \in B^{-1} (Y)} \! P (a | X) \label{CTC}
\end{align}
\vspace{-1.2em}

\noindent 
and under the independence assumption:
\vspace{-0.5em}
\begin{align}
	P (a | X) = \prod_{i=1}^{T} P (a_i | X)
\end{align}
\vspace{-1.2em}


\noindent
Equation \ref{CTC} can be efficiently computed using a forward-backward dynamic programming method as described in \citet{graves2006connectionist}. Decoding (finding the most likely $Y$ from the output sequence $X$) can be done by performing a beam search \citep{graves2014towards}, or simply by selecting the single most likely prediction at each timestep and then applying the mapping function $B$:
\vspace{-0.5em}
\begin{align}
\argmax_Y P(Y | X) \approx B(\argmax_a P(a | X))
\end{align}
\vspace{-1.2em}

\vspace{-0.5em}
\subsection{CTC Training with Per-Timestep Supervision}
\begin{figure}[h!]
\centering
\includegraphics[width=0.99\linewidth]{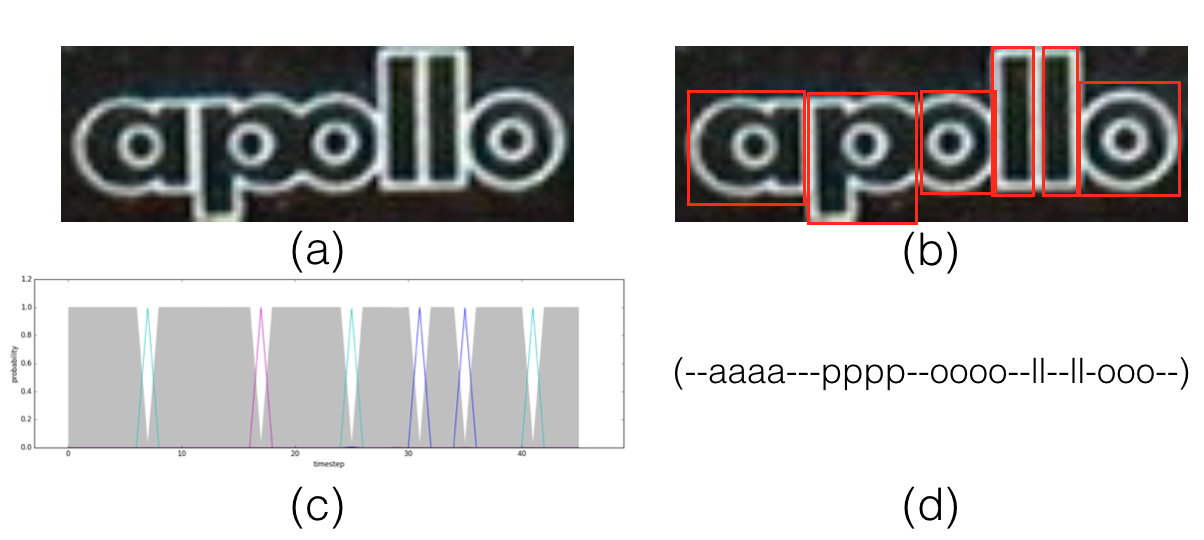} 
\vspace{-1.5em}
\caption{(a) A word image in the training dataset. (b) A word image with ground-truth character-level bounding boxes. (c) A typical output sequence from a CTC-trained model. The gray area corresponds to the probability of outputting a blank label at a given time-step, while colors indicate non-blank labels. Each non-blank label occurs as an isolated peak. (d) A per-timestep groundtruth generated based on (b).}
\vspace{-0.5em}
\label{fig:blanks}
\end{figure}

During CTC training process, blank labels typically dominant the output sequence. Non-blank labels only appear as isolated peaks, an example of which is depicted in Figure \ref{fig:blanks}. This is a consequence of the forward-backward algorithm  \citep{graves2006connectionist}. Since we add a blank label between each character, there are more possible paths going through a blank label at a given timestep in the CTC forward-backward graph. In the early stage of CTC training where model weights are randomly initialized, all paths have similar probabilities. As a result, the probability of a given timestep being a blank label is much higher than any of the others when summing all valid alignments (paths in CTC graph) in Equation \ref{CTC}. Furthermore, the back-propagated gradient computed using the forward-backward algorithm will encourage blank labels to occur at each single time-step.

Predicting only blanks will lead to a high CTC loss and, as a result, the model will try to predict characters at specific timesteps. However, due to the blank label issue described above, it generally takes many iterations before a non-blank label appears in the output sequence during CTC training. To accelerate the training process, we introduce per-timestep supervision. If we have access to the character-level bounding boxes in each word image, then we would be able to decide the label of $x_i$ at a timestep $i$, based on its corresponding receptive field. In our experiments, $x_i$ is assigned a label $z_i = y_j$ if its receptive field overlaps with more than half of the area of $y_j$, otherwise $x_i$ is assigned a blank label. During training, the objective function becomes:
\vspace{-0.5em}
\begin{align}
	L(X) &= CTC(X) + \lambda L_{pt}(X)\\
	L_{pt}(X) &= \frac{1}{T}\sum_{i=1}^{T} - \text{log} P(z_i | x_i)
\end{align}
\vspace{-1.2em}

\noindent where $\lambda$ is a hyper-parameter meant to balance the two terms in the loss. Since our per-timestep supervision only provides one possible alignment, it does not guarantee exact solution to Equation \ref{CTC}. Thus we decrease $\lambda$ throughout training. At the start $\lambda$ is set such that the gradients yielded by two kinds of losses have same magnitude.

At the time this paper was being written, there was no large-scale public dataset that provides character-level bounding box information, so we generated our own synthetic data similar to  \citet{jaderberg2014synthetic}, except that we kept tracking the location of each character. Such dataset will be made public for further study.

\vspace{-0.2em}
\subsection{CTC Training with Decoding Penalty}
Another issue of CTC training is the gap between the objective function and the evaluation criterion. As stated in \citet{graves2014towards}, CTC loss will try to maximizes the log probability of outputting \textit{completely correct} label $Y = \{y_1, y_2, \cdots, y_L\}$. In another word, incorrect predictions are treated as equally bad. However, this is not always the way the model performance is assessed. For example for scene text recognition task, people often report edit distance between the prediction and the groundtruth. \citet{graves2014towards} proposed a sample-based method to calculate the expected loss. However the involved sampling step would slow down training process.

Here we proposed a simpler solution to penalize bad predictions. A revised version of CTC is introduced:
\vspace{-0.5em}
\begin{align}
	WCTC(x) = - \text{log} P (Y | X) \cdot L_e (Y, Y_D)
\end{align}
\vspace{-1.2em}

\noindent
where $L_e (\cdot, \ \cdot) $ is a real-value loss function (e.g. edit distance between two strings) and $Y_D$ is the decoded prediction. Depending on the decoding method performed, $Y_D$ could be different. For example, one can approximate $Y_D$ by:
\vspace{-0.5em}
\begin{align}
Y_D = B(\argmax_a P(a | X)) \approx \argmax_Y P(Y | X)
\end{align}
\vspace{-1.2em}

$WCTC(X)$ would inevitably emphasize longer $Y$, therefore we suggest that $WCTC(X)$ should step up after several epoches of CTC training, where most $Y_D$ have reasonably small edit distance compared with $Y$.

\subsection{Text Correction}\label{sec:correction}

The text recognition model of our system expects cropped image patch containing single word as input. However, this may not be always satisfied in the text localization step. The generated bounding box is sometimes incomplete, resulting in word recognition with missing characters. Moreover, spaces between words might be hard to distinguished from spaces between characters. Consequently, a post-processing step is necessary to correct the outputs of the recognition model. During experiments, we found that a standard spell-checker is not powerful enough for automatic correction, since we need to not only correct individual words, but also break larger strings into words. Here we train another RNN to address this issue, employing a character-level sequence-to-sequence model \citep{sutskever2014sequence}. 

\begin{figure}[htb!]
\centering
\vspace{-0.3em}
\includegraphics[width=0.3\linewidth, height=3em]{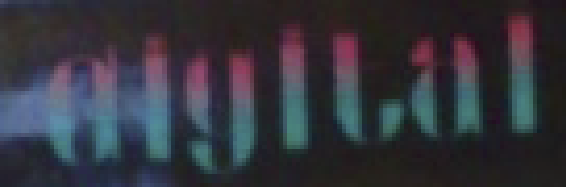} 
\includegraphics[width=0.6\linewidth, height=3em]{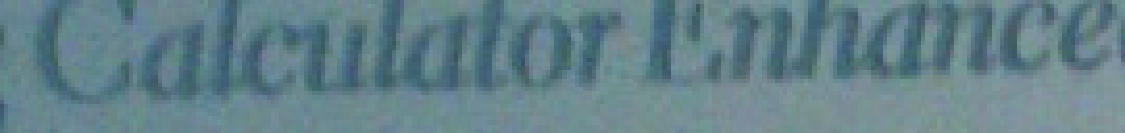}
\includegraphics[width=0.45\linewidth, height=3em]{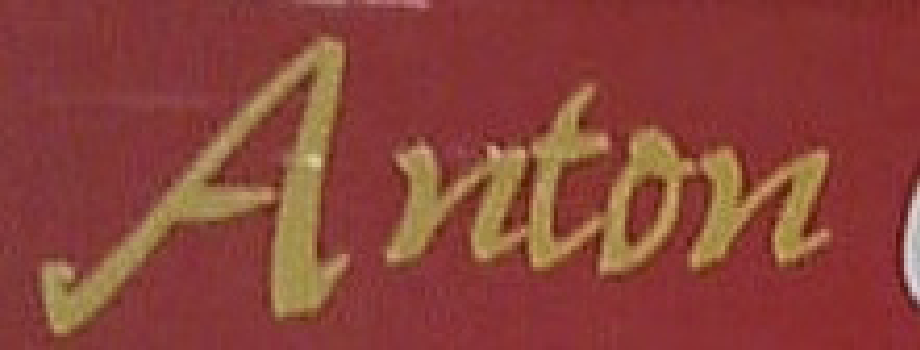}
\includegraphics[width=0.45\linewidth, height=3em]{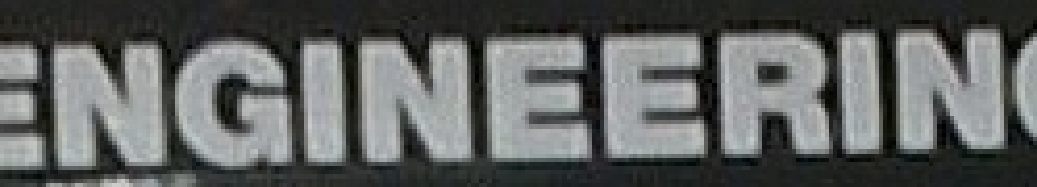}
\vspace{-0.5em}
\caption{Text recognition examples generated by Tesseract \citep{smith2007overview}, Our recognition model and correction model. From top to bottom, left to right: (a) Tesseract: mum, fUdlCBldUl ItnMnCe, Ahmn, ELIGINEERIM (b) Our recognition: oigital, CalculatorEnnance, Anton, NGINEERIN (c) Our correction: digital, Calculator Enhance, Anton, ENGINEERING.}
\label{figure:correction}
\end{figure}

Given an input sentence with random noise (deletion, insertion, or replacement of randomly selected characters), the sequence-to-sequence RNN is expected to output correct one. The RNN is designed with $2$ layers of LSTM, each with a hidden size of $512$. During training, sampled sentences from the Google one-billion-word dataset \citep{chelba2013one} and their noised variants are used as targets and inputs, respectively. Figure \ref{figure:correction} illustrates several cropped image samples which demonstrate the effectiveness of our correction model.

Our recognition model is trained in an end-to-end manner, while \citet{he2016reading} trained CNN part and RNN part separately. Both our recognition model and correction model is trained using mini-batch stochastic gradient descent (SGD) together with Adadelta \citep{zeiler2012adadelta}.


\vspace{-2.2em}
\section{Experiments}
\label{sec:experiment}

\subsection{Text Recognition}

To assess the performance of our proposed text recognition method, we report our results on several benchmark datasets: ICDAR 2003 (IC03), SVT and III5K following the standard evaluation protocol in \citet{wang2011end}. IC03 provides lexicons of 50 words per image (IC03-50) and all words of the test set (IC03-Full). Images in SVT and III5K are also associated with lexicons (SVT-50, III5K-50, III5K-1K).

We refer to our base model, trained using the CTC loss, as the Deep Sequential Labeling model (DSL), while the model trained using the revised loss functions, $CTC(X) + \lambda L_{pt}$ and $WCTC(X) + \lambda L_{pt}$, will be referred to as DSL-CL and DSL-WL respectively.

\begin{figure}[h!]
\centering
\vspace{-0.5em}
\includegraphics[width=0.49\linewidth]{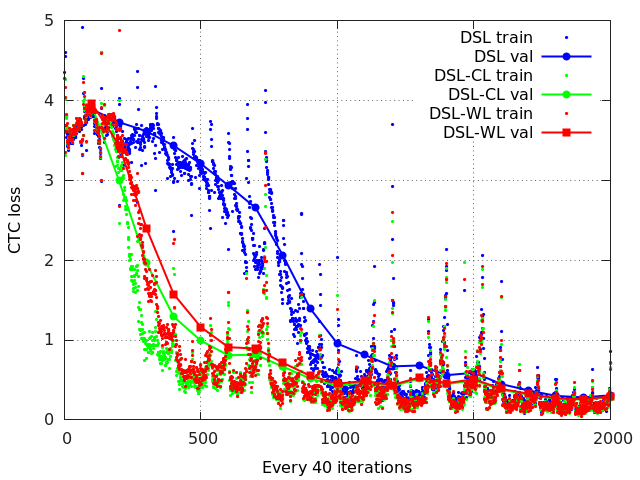} 
\includegraphics[width=0.49\linewidth]{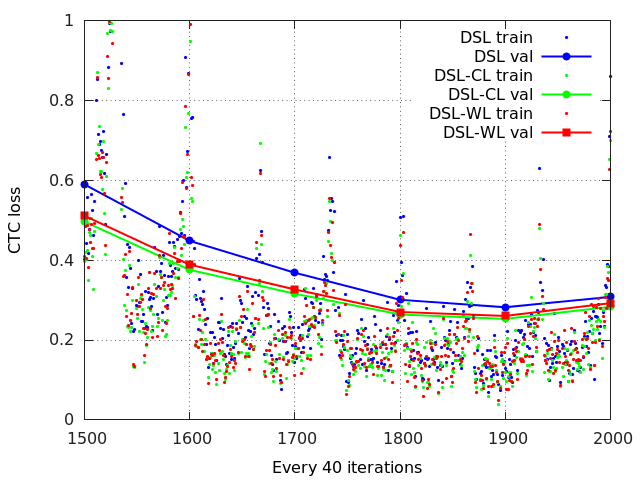}
\vspace{-0.6em}
\caption{Left: CTC loss during training using different objective functions. The dots are the training loss and the solid lines are the validation loss. Right: a zoomed-in version at a later stage.}
\vspace{-1em}
\label{figure:res:training}
\end{figure}

Figure \ref{figure:res:training} shows the CTC loss during training using different objective functions. To perform efficient mini-batch training, we group data samples based on the length of their $Y$. Therefore one can observe sudden jumps in the loss curve, which corresponds to longer $Y$ in that batch. As we can see, adding per-timestep classification loss would significantly speedup training at early stage. At later stage, as $\lambda$ becomes smaller and smaller, the difference between with and without $L_{pt}$ decreases. However, from validation loss we can still observe certain benefits when using $L_{pt}$. 

Table \ref{table:res:text} shows the results of the trained recognition model compared to Tesseract (Smith 2007) used in \citet{tsai2011combining} as well as some other state-of-the-art methods. We can see that Tesseract performs poorly in all cases. It is clear that distortions in real scene images make recognition a very different and challenging task from a standard OCR system. Our recognition models outperforms methods with handcrafted-features \citep{wang2011end,lee2014region,yao2014strokelets} and several methods that make use of deep neural representations for individual characters \citep{wang2012end,jaderberg2014deep}, indicating that our model yields much better results for recognition by learning sequential information. Our base model DSL produces better results on all datasets except IC03-50 compared with \citet{he2016reading}. We attribute this to the end-to-end training procedure and large amount of synthetic data during training. Jaderberg 2015b model achieves the best results for IC03-50 and IC03-Full. However, since they treat text recognition problem as a multi-class (number of classes equals number of words) classification task, it is difficult for their model to train and to adapt to out-of-dictionary text. Our base model DSL is expected to achieve similar performance to Shi et al. 2016 since we share similar model architecture and training procedure. We attribute the difference to the different synthetic training data in use and the hyper-parameter tuning.

By employing a per-timestep loss $L_{pt}$, our model would no longer output isolated peaks of non-blank prediction. Moreover, from Figure \ref{figure:res:training} and Table \ref{table:res:text} we can see that DSL-CL and DSL-WL performs better than DSL. We hypothesize that peak prediction is harmful since it would confuse the model about where to yield non-blank prediction and where not to. 

DSL-WL ties or slightly outperforms DSL-CL in all datasets, suggesting that our revised loss $WCTC(X)$ is effective.

\begin{table}[h!]
\centering
\vspace{-0.5em}
\begin{tabular}{ | l | p{20pt} | p{20pt} | p{20pt} | p{20pt} | p{20pt} |} 
 \hline
 \multirow{2}{2em}{Method} & \multicolumn{5}{ c |}{Recognition Accuracy(\%)}\\ \cline{2-6}
 & IC03-50 & IC03-Full & SVT-50 & III5K-50 & III5K-1K\\ \hline
 Smith 2007 & 60.1 & 57.3 & 65.9 & - & -\\ \hline
 Wang et al. 2011 & 76.0 & 62.0 & 57.0 & 64.1 & 57.5\\
 Lee et al. 2014 & 88.0 & 76.0 & 80.0 & - & - \\
 Yao et al. 2014 & 88.5 & 80.3 & 75.9 & 80.2 & 69.3\\ \hline
 Wang et al. 2012 & 90.0 & 84.0 & 70.0 & - & - \\
 Jaderberg 2014 & 96.2 & 91.5 & 86.1 & - & -\\
 Jaderberg 2015b & \textbf{98.7} & \textbf{98.6} & 95.4 & 97.1 & 92.7\\ \hline
 Shi et al. 2016 & \textbf{98.7} & 97.6 & \textbf{96.4} & 97.6 & 94.4\\ 
 He et al. 2016 & 97.0 & 93.8 & 93.5 & 94.0 & 91.5\\
 Our DSL & 96.1 & 94.7 & 94.5 & 97.7 & 95.1\\ 
 Our DSL-CL & 96.6 & 95.3 & 94.5 & 98.3 & 95.9\\ 
 Our DSL-WL & 98.2 & 95.8 & 94.6 & \textbf{98.5} & \textbf{96.0}\\ \hline
\end{tabular}
\vspace{-0.3em}
\caption{Cropped word recognition accuracy across several benchmark datasets.}
\vspace{-1.5em}
\label{table:res:text}
\end{table}

\subsection{Book Spine Retrieval}

To assess the retrieval performance of our system, we adopt the retrieval-based evaluation method similar to that of \citet{chen2010building} and \citet{tsai2011combining}. However, since we only have access to the $454$ book spine images that they used for querying instead of the whole database to search from, which contains $2,300$ books, it is necessary to build our own collection. Based on the topics of those $454$ books, we crawled and sampled over $9,000$ books from the Stanford Engineering Library under the subjects of computer science, electrical engineering, mathematics, etc. to construct our database, together with those $454$ test spine images. For each book in our database, we crawled its title and meta-data such as the author and publisher. This information is then indexed for later retrieval tasks. We expect that the dataset we built is a superset of theirs, which would mean that higher precision and recall from our results could indicate superior performance.

Our system only relies on text spotting. As a result, we only need the textual information of each book and do not require storing and querying spine images, dramatically reducing the volume of data we need to store and manage. Moreover, for libraries with limited resources, one cannot always assume that book images are already available. 

For each book spine image, text is detected, recognized, and corrected. The outputs are further refined by matching to a dictionary from our database using a nearest neighbor matching algorithm. Finally, we use these outputs as keywords to search from our database. During search, \emph{tf-idf} (term frequency-inverse document frequency) weights are used to rank returned results. The top-ranked result, if it exists, is declared as our prediction for each book spine image. We built our search engine based on Apache Solr so that our system scale well to large collection of books.

Much as in \citet{chen2010building} and \citet{tsai2011combining}, we report the precision and recall when querying the $454$ spine images. The precision is defined as the proportion of correctly identified books to the declared correct books, while the recall is defined as the proportion of correctly identified books to all queries. We further defined recall at top-$k$, which measures the number of correctly identified books to appear in top-$k$ results of a search. Since only one book spine in our database could be correct for each query, precision at top-$k$ is not meaningful as it will linearly decrease against $k$.  We also evaluate our results using  Reciprocal Rank ($RR$) which is defined as: $RR = \sfrac{1}{K}$, where $K$ is the rank position of the first relevant document (the target book spine in our case). Average Reciprocal Rank ($MRR$) across multiple queries is reported. All these measures are widely used by the information retrieval community.

Using top-$1$ search results, our system yields a $MRR$ of $0.91$. Table \ref{table:res:compare} shows our results compared with several other methods. The results of \citep{tsai2011combining} are extracted from precision/recall curves in the paper which yields best F-score. Using only textual information, we achieve the best recall and reach a higher (0.91 VS 0.72) F-score. Given that our database is much larger than theirs (over $9,000$ versus $2,300$), the results show the superiority of the proposed. \citet{tsai2011combining} achieved 0.92, 0.60 and 0.91 for precision, recall and F-score respectively in their Hybrid model, however the hybrid model relies on both text and image querying, which is time-consuming.  

\begin{table}[h!]
\centering
\vspace{-0.5em}
\begin{tabular}{ | l | c | c | c | } 
    \hline
    & Precision & Recall & F-score\\ \hline \hline
    Tsai2011 (Text) & \textbf{0.92} & 0.60 & 0.72 \\
    Ours & \textbf{0.92} & \textbf{0.90} & \textbf{0.91} \\ \hline
\end{tabular}
\vspace{-0.5em}
\caption{The precision and recall at top $1$ compared with other methods, using only text as queries. Note that our precision and recall come from a much larger database from which to search books.}
\vspace{-1em}
\label{table:res:compare}
\end{table}

\begin{figure}[h!]
\centering
\includegraphics[width=0.6\linewidth]{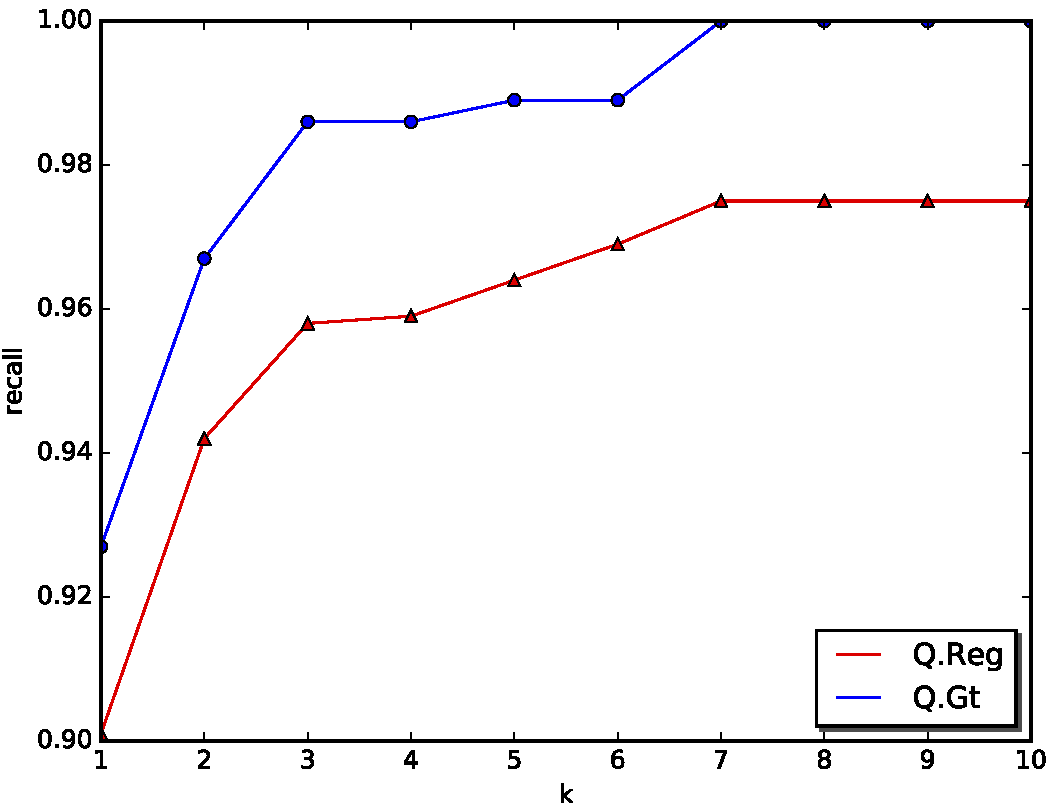} 
\vspace{-1.2em}
\caption{The recall at top-$k$ during retrieval. Q.Reg means that we use recognized text as our keywords during search, while Q.Gt means that we use groundtruth titles (no meta-information) as keywords.}
\vspace{-1.8em}
\label{figure:res:recall}
\end{figure}

Figure \ref{figure:res:recall} shows the recall at different rank positions $k$. Our model achieves $96.4\%$ recall within top $5$ search results, demonstrating the effectiveness of our text spotting system. Note that since our book collection contains books with very similar titles, even using groundtruth titles as keywords in search cannot guarantee $100\%$ recall at top-$1$ rank position.

When further investigating the failure cases, we found that a large portion of wrong predictions were due to less discriminative keywords used during search. Multiple books may have very similar or even identical titles, with minor differences in meta-information such as author name, publisher name, volume number, etc. Moreover, some meta-information on book spine images tends to be blurry and small in size, making detection and recognition more difficult. In such cases, employing a text-based search might be inadequate. To develop a more quantitative understanding how these factors might diminish the performance of our model, we evaluate the retrieval performance by querying ground-truth titles on book spine images. Results are shown in Figure \ref{figure:res:recall}. These results provide a reasonable idea of the potential upper bound on the performance of a text-based method. Although image-based search might address this issue, it comes with the steep cost of storing and transmitting images. The rest of the failure cases are majorly due to imperfect or even wrong text localization bounding boxes.

\vspace{-0.2em}
\section{Conclusion}
\label{sec:conclusion}
We propose a system which uses a state-of-the-art deep neural architecture for scene text detection and recognition for the purpose of identifying specific books in the library and managing library's inventory efficiently. We achieve state-of-the-art performance for scene text recognition task on several benchmark dataset while reducing training time. An information retrieval experiment is conducted, using large physical library database, to evaluate the performance of the whole system. We demonstrate that text recognition is competitive with image-matching retrieval, while text recognition based retrieval reduces the need for storing or transmitting book spine images, which may not be available to all users.

\clearpage
\bibliographystyle{aaai}
\bibliography{aaai}
\end{document}